\documentclass[runningheads]{llncs} 
\author{Zeeshan Khan Suri\orcidID{0000-0001-9036-3195}
}
\authorrunning{Zeeshan Khan Suri}
\institute{DENSO ADAS Engineering Services GmbH, Lindau 88131, Germany
\email{z.suri@eu.denso.com}\\
}

\title{Pose Constraints for Consistent Self-supervised Monocular Depth and Ego-motion}
\titlerunning{Pose Constraints for Consistent Self-supervised Monocular Depth}

\usepackage{xspace}

\makeatletter
\DeclareRobustCommand\onedot{\futurelet\@let@token\@onedot}
\def\@onedot{\ifx\@let@token.\else.\null\fi\xspace}

\def\ie{\emph{i.e}\onedot}

\def\wrt{w.r.t\onedot} 
\def\etal{\emph{et al}\onedot}
\makeatother

\usepackage{graphicx}
\usepackage{multirow}
\usepackage{amsmath}
\usepackage{amssymb}
\usepackage{booktabs}
\usepackage[inline]{enumitem}
\RequirePackage[table]{xcolor}
\RequirePackage{graphicx,subfigure}

\usepackage[capitalize]{cleveref}
\crefname{section}{Sec.}{Secs.}
\Crefname{section}{Section}{Sections}
\Crefname{table}{Table}{Tables}
\crefname{table}{Tab.}{Tabs.}
\begin{document}

\maketitle

\begin{abstract}
Self-supervised monocular depth estimation approaches suffer not only from scale ambiguity but also infer temporally inconsistent depth maps \wrt scale. While disambiguating scale during training is not possible without some kind of ground truth supervision, having scale consistent depth predictions would make it possible to calculate scale once during inference as a post-processing step and use it over-time. With this as a goal, a set of temporal consistency losses that minimize pose inconsistencies over time are introduced. Evaluations show that introducing these constraints not only reduces depth inconsistencies but also improves the baseline performance of depth and ego-motion prediction.
\end{abstract}

\keywords{self-supervision \and 3d-reconstruction \and depth-estimation}


\section{Introduction}
\label{sec:intro}
Depth is so essential for perceiving, understanding and navigating the 3D world around us that biological animals have evolved to possess redundant apparatus for perceiving it. Animals are able to infer relative depth of the perceived scene even with a single eye \cite{monocularrats}. Although the loss of dimensionality from projecting a 3D scene on a 2D plane cannot be completely recovered, an image consists of many useful monocular cues, such as relative sizes, texture gradient, linear perspective, contrast differences. Image sequences contain additional motion cues such as occlusion, motion parallax \cite{opticsofeuclid}. These cues impose constraints on the possible combinations of depth of the underlying scene. Such cues can be implicitly learnt by computational models in a supervised \cite{make3d,dorn}, and in a self-supervised \cite{sfmlearner,monodepth2,packnet} manner, making it possible to infer relative depth from a single image. During training, the self-supervised methods rely on motion cues coming from image sequences. They simultaneously estimate relative pose between two consecutive frames of a sequence and their individual depth such that by warping one frame onto the other using the depth, relative pose and camera intrinsics, the other frame can be synthesized. The underlying assumption is of photometric constancy of neighboring frames of a sequence, and thus a loss between the warped neighboring image and the true one can back-propagate through the depth and pose networks.

\begin{figure}[hbt!]
\centering
\begin{tabular}{cc}
\includegraphics[width=0.47\textwidth]{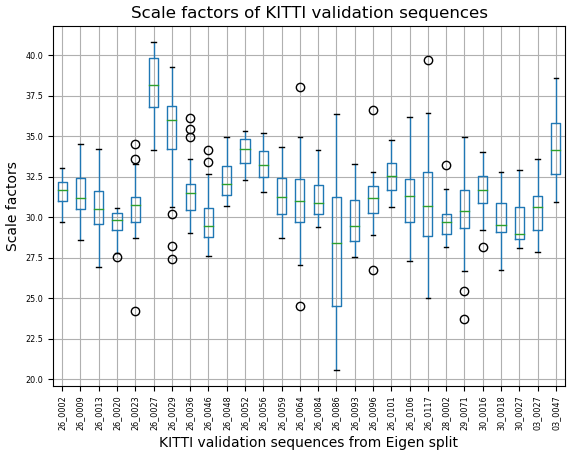} &
\includegraphics[width=0.53\textwidth]{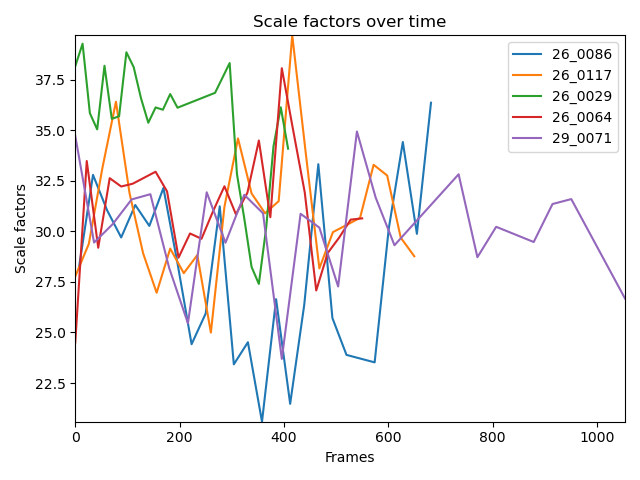}\\
\end{tabular}
  \caption{Variation of scale factors $\frac{\text{median}(D_{\text{true}})}{\text{median}(D_{\text{pred}})}$ per frame, within each KITTI validation sequence is shown (\textbf{Left}): as a box-plot, and (\textbf{Right}): over time.}
  \label{fig:md2scales}
\end{figure}

The 2D image pixels might have been projected from infinite number of 3D points, thus making the inverse problem of recovering the depth dimension from the 2D image ill-posed. While cues constrain the solution depth map to have an underlying structure, the estimated depth can only be relative in nature and not absolute, since any scalar multiple of the estimated depth map could also be equally optimal. Furthermore, these self-supervised models do not necessarily learn our standard units of measuring distance, but rather predict depth in their own. Even for humans these units, such as metric, imperial, US customary, tend to vary in usage, for, after all, these units and the scalar factors relating them are a human construct. Such scale ambiguity is not usually a problem, since one can calculate the scale factor relating the model's depth to that of metric or imperial as a post-processing step and convert one to the other. Unfortunately, the depth estimates from self-supervised models are not just scale ambiguous but also temporally scale inconsistent, \ie the scale of one frame's depth map is different to that of the neighboring frame's. This variation of the scale factors is shown in \cref{fig:md2scales} as a box-plot for each KITTI validation sequence and also over-time for the top 5 varying sequences. Here, the scale factor of a frame is the ratio of the metric LIDAR depth to that of the predicted depth $=\frac{\text{median}(D_{\text{true}})}{\text{median}(D_{\text{pred}})}$.

\vspace{-0.5em}
\vspace{-0.5em}

Due to this variation, it is not accurate to analytically determine the scale factor once and use it later, and makes it necessary to calculate the scale in each frame, making it infeasible for applications without availability of some kind of ground-truth, as in case of online videos. In fact, most current methods artificially scale their outputs depth maps per-frame, using the LIDAR ground truth, as a post-processing step before calculating the error at test time. 

This work hypothesises that the scale consistency problem is due to the lack of proper temporal constraints, \ie the scale is independently ambiguous for each frame, and the training pipeline finds a scale that is locally optimal for that frame, regardless if it agrees with the scale of the neighboring frames. While some of the recent works on this issue impose consistency in depth, this work instead explores additional ways of imposing temporal constraints in an unsupervised manner, \ie without the need of additional ground truth supervision.


\section{Related Work}

This section explores different ways the scale ambiguity and the scale consistency problem has been tackled in the literature. Also losses similar to ours in related tasks are referred. Temporal data such as video has temporal correlations and has a low probability to abruptly change within short intervals. Temporal consistency exploits this information flow and has been utilized for various video applications \cite{temporalconsistency12,temporalconsistency15,temporalconsistency20}, including supervised depth estimation \cite{tcsuperviseddepth19}, style transfer \cite{videostyletransfer17}, video completion \cite{tcvideocomplete16} segmentation \cite{tcsegmentation22} to capture temporal correlations. Temporal consistency is also the key component of the recent self-supervised monocular depth estimation approaches \cite{depthtransfer14,sfmlearner}, where the photometric constancy among consecutive frames is assumed and differences after reprojection are minimized. But, since the depth network in these methods is monocular and sees only one frame at a time, the predicted depth maps are not consistent over time.

\subsection{Scale-Disambiguation/Consistency-Enforcement via Supervision}
Methods with some kind of ground truth supervision are able to enforce scale consistency and do not have the same scale inconsistency issues, for example, the scale factor in stereo methods comes from the relation of disparity and depth as $\text{depth}=\frac{\text{focal length} \times \text{baseline}}{\text{disparity}}$, and is constant. Following this, Roussel \etal \cite{stereo_scaling19} first pre-train on stereo data and then fine-tune on monocular data, to show that doing so retains the scale learnt from the stereo pretraining. GPS or ground-truth pose data has also been used to disambiguate scale \cite{packnet,chawlavarma2021multimodal} via enforcing the pose network's output to match the pose coming from sensors.

Hand-picked features based depth estimation methods such as SLAM \cite{orbslam2} and Structure from Motion (SfM) \cite{sfmrevisited16} have been used \cite{prgbdslam20,consistentvideodepth20} for a source of supervision, to transfer wide baseline symmetric depth and sparse long-term depth consistencies from it to the depth estimation network. In a similar fashion, ideas from the Visual Odometry (VO), such as epipolar geometry and bundle adjustment, are incorporated \cite{scalevo20,dfvo20}, to independently compute correspondences and triangulate them to produce sparse depth, which acts as additional supervision to the network's depth prediction. 

\textbf{Via the Plane Assumption.}
Following Kitt \etal \cite{kittmonocular11}'s scale recovery for VO, many recent works \cite{planecalib20,planetowards20} make use of the following assumptions:
\begin{enumerate*}[label=\alph*)]
 \item most automotive cameras are rigidly mounted in a fixed position and at constant orientation with respect to the road,
 \item the roll and pitch movement of the vehicle have negligible effect on its position and orientation,
 \item most urban streets may be assumed to be approximately planar in the vicinity of the vehicle
\end{enumerate*}. These assumptions allow them to use camera extrinsic parameters, in particular the camera height and compare it with the estimated height by fitting plane on the road, to recover the scale as a post-processing step. \cite{planeseg21,plane22} also use camera height for scale but incorporate it within training. These methods rely on heuristics that a flat road plane is visible in the area of interest and that the camera position and orientation remain constant over time, which are often not realistic.

Methods that use some kind of supervision for disambiguating depth at each frame, also inherently enforce temporal consistency. Since the supervision is temporally consistent, the predicted depth maps are also made temporally consistent as an unintended consequence. 

\subsection{Self-supervised Temporal Consistency}
Recurrent neural networks \cite{wang2019recurrent,lstmdepth20} have been used to implicitly model multiple frames inputs. Having multiple frames as input directly to the depth network causes holes at regions with moving objects \cite{manydepth}, and need to be corrected by an additional single input network.
3D geometric constraints \cite{vid2depth,3dconsistency19,3dscale21} were proposed that penalize the euclidean distances between the reconstructed point clouds of two consecutive frames, after transforming one to the other. Similar geometric constraints on the depth maps were proposed \cite{scsfmlearner21,tcdepthsupervised21,tcdepth21} which minimize the inconsistency of the estimated disparity maps of two consecutive frames, after warping one onto the other. Our work lies in this category and differs in the fact that we propose complementary constraints on the pose, which can, in principle, be added and used together with the other temporal constraints.

\subsection{Similar Constraints in Literature}
Constraints similar to the ones we impose via a loss are found across various computer vision applications.

\textbf{Forward-Backward Consistency.}
Li \etal \cite{undeepvo18} propose a forward-backward pose consistency loss but in a stereo setting, where, the pose from one frame of the left camera to its neighboring frame should be identical to the pose between the same frame of the right camera to its neighboring frame. Based on Narayanan \etal \cite{offorwardbackward10}'s forward-backward optical flow consistency assumption, \cite{unflow18} propose a loss in their optical flow prediction network. This was adopted in the context of monocular depth and ego-motion \cite{geonet,wang2019recurrent}, where the flow caused by rigid ego-motion estimation is computed and the forward-backward inconsistency of the rigid flow is minimized. Sheng \etal \cite{keyframecyclewarp21} propose a loss that minimizes the forward-backward inconsistencies in the bi-directional warping fields generated from the rigid ego-motion. \cite{dfnet18,dfvo20} train an optical flow network, in addition to the depth and ego-motion networks for per-pixel dense 2D pose between two consecutive frames and adopt the same forward-backward consistency loss for their optical flow. Li \etal \cite{li2021unsupervised} propose a forward-backward loss directly on pose estimates. Our forward-backward loss is similar to theirs.

\textbf{Cycle Consistency.}
The forward-backward consistency is an instance of the idea of cycle-consistency which has been applied to a wide range of computer vision tasks \cite{cycleshapematch13,cyclecorr19,cycledomainada18,cyclegan20}. Related to our problem of interest, pose cycle consistency was used in the context of Visual Odometry \cite{cyclevo18,lstmvotc20}. Ruhkamp \etal \cite{tcdepth21} propose a strategy to detect and mask inconsistent regions such as occlusions, in neighboring depth maps via a cycle consistency of the reprojected RGB frames. 

\section{Method}
Given video sequences captured from a camera with known intrinsic parameters $K$, the objective is to learn a depth network model $f_D: \mathbb{R}^{3\times H\times W} \rightarrow \mathbb{R}^{H\times W}$ that maps an RGB frame at time $t$, $I_t\in \mathbb{R}^{3\times H\times W}$ to the depth of its underlying scene $D_t\in \mathbb{R}^{H\times W}$, and an ego-motion network model $f_E: \mathbb{R}^{2(3\times H\times W)} \rightarrow \mathfrak{se}(3)$ that takes in two consecutive RGB frames $\{I_t, I_{t+n},\,$ typically, $n \in \{-1,1\}\}$ and outputs the 6 Degrees-of-Freedom (DOF) rigid transformation $\mathbf{T}_{t}^{t+n} = (\mathbf{e}^T, \mathbf{t}^T)^T \in \mathfrak{se}(3), \text{ and}, \big(\begin{smallmatrix}
 \mathbf{R} & \mathbf{t}\\
 0 & 1
\end{smallmatrix}\big) \in \text{SE}(3)$ from $t$ to $t+n$, where $H,W$ are the image height and width and, $\mathbf{e} \in \mathfrak{so}(3), \mathbf{t} \in \mathbb{R}^3, \mathbf{R} \in \text{SO}(3) $ represent the axis-angle, translation and the rotation matrix respectively.

The depth and the ego-motion are jointly trained with the key assumption of photometric constancy, \ie a view reprojected with the camera intrinsics, depth and the camera motion onto its neighboring view, reconstructs the neighboring view. The reprojection of a view onto the neighbor is as follows
\begin{equation}\label{eq:reprojection}
[\hat{u}, \hat{v}, 1]^T \sim K\mathbf{R}_t^{t+n}D_t^{ij}K^{-1}[u,v,1]^T+K\mathbf{t}_t^{t+n},
\end{equation}
where $[u,v,1]$ and $[\hat{u}, \hat{v}, 1]$ denote the homogeneous pixel coordinates of view at time $t$ and that of its neighboring view, $D_t^{ij}$ is the pixel's corresponding depth and $\mathbf{R}_t^{t+n}, \mathbf{t}_t^{t+n}$ denote the rotation and translation from the view at time $t$ to that of the neighboring view at time $t+n$ respectively. Using the reprojection \cref{eq:reprojection}, the view at time t is synthesized from the neighboring view as
\begin{equation}\label{eq:viewsynthesis}
\hat{I}_{t+n\rightarrow t}[u,v] = I_{t+n}\langle [\hat{u}, \hat{v}] \rangle,
\end{equation}
where $\langle\rangle$ denotes the sampling operator. A robust photometric loss \cite{photometricloss16,monodepth17} between the synthesized RGB view $\hat{I}_{t+n\rightarrow t}$ to the original one $I_{t}$, as in \cref{eq:viewsynthesis}, is minimized, thereby updating and correcting the depth and ego-motions model weights via backpropagation.

\begin{equation}
  \mathcal{L}_p(I_{t}, \hat{I}_{t+n\rightarrow t})= \frac{\alpha}{2}(1-\text{SSIM}(I_{t}, \hat{I}_{t+n\rightarrow t})) + (1-\alpha)\lVert I_{t} - \hat{I}_{t+n\rightarrow t}\rVert_1,
\end{equation}
where $\alpha=0.85$ and SSIM denotes the structural similarity loss \cite{ssim}. Following Godard \etal \cite{monodepth17} an additional edge $\delta_x, \delta_y$ aware smoothness term regularizes the predicted depth:
\begin{equation}
  \mathcal{L}_s= \sum_{uv}\lvert \delta_x \bar{D}_t^{uv} \rvert e^{-\lvert\delta_x I_t^{uv}\rvert} + \lvert \delta_y \bar{D}_t^{uv} \rvert e^{-\lvert\delta_y I_t^{uv}\rvert},
\end{equation}
where $\bar{D}^{uv}_t$ is the mean normalized depth at pixel $[u,v]$.
We establish a strong baseline by following the best practices from Monodepth2 \cite{monodepth2}. We also use the minimum reprojection error $\min_n \mathcal{L}_p(I_{t}, \hat{I}_{t+n\rightarrow t})$, over pairs of both neighboring frames to deal with occlusions and auto-masking to disregard temporally static pixels. These losses are minimized over all pixels in the training set over four resolution scales. The readers are referred to Monodepth2 \cite{monodepth2} for more details.

\subsection{Pose Constraints}

The depth and the pose networks are tightly coupled. 
We propose to establish temporal consistency in the predicted depth through the pose network's estimates of the ego-motion between two input frames. Specifically, we propose three constraints that do not add any additional assumptions, but are expected to be met explicitly. Let $T = \big(\begin{smallmatrix}
 \mathbf{R}_T & \mathbf{t}_T\\
 0 & 1
\end{smallmatrix}\big) \in \text{SE}(3)$ and $P = \big(\begin{smallmatrix}
 \mathbf{R}_P & \mathbf{t}_P\\
 0 & 1
\end{smallmatrix}\big)\in \text{SE}(3)$ be two 6-DOF rigid poses that represent the same transformation, we define the distance metric $d(T,P): \text{SE}(3) \times \text{SE}(3) \rightarrow \mathbb{R}^+$, between them as 


\begin{equation}\label{eq:se3dist}
  \mathit{d}(T,P) = d(\mathbf{R}_T, \mathbf{R}_P) + d(\mathbf{t}_T, \mathbf{t}_P) = \Big\lVert 1-\left(\frac{\text{tr}(\mathbf{R}_T\mathbf{R}^T_P) - 1}{2}\right) \Big\rVert_1 + \lVert \mathbf{t}_T - \mathbf{t}_P \rVert_1
\end{equation}
, where $\theta = \cos^{-1} (\frac{\text{tr}(\mathbf{R}_T\mathbf{R}^T_P) - 1}{2}) \in [0,\pi]$ is the angle of the relative rotation $\mathbf{R}_T\mathbf{R}^T_P$, and $1-\cos{(\theta)}$ corresponds to the geodesic distance on the 3D manifold of rotation matrices \cite{metricsforrotations09,angle_rots}. Throughout the proposed constraints, we use \cref{eq:se3dist} as the distance metric between SE(3) transformations.

\subsubsection{Forward-Backward Pose Consistency.}
The ego-motion between frames at time $t$ and $t+n$ should be the inverse of the ego-motion between frames at time $t+n$ and $t$, \ie $\mathbf{T}^t_{t+n} \stackrel{!}{=} {\mathbf{T}_t^{t+n}}^{-1}$. This is characterized by minimizing the following loss
\begin{equation}
\mathcal{L}_{\mathbf{T}_{fb}} = d(\mathbf{T}^t_{t+n}, {\mathbf{T}_t^{t+n}}^{-1}),
\end{equation}
, where $\mathbf{T}^{-1} = \big(\begin{smallmatrix}
 \mathbf{R}^T & -\mathbf{t}\\
 0 & 1
\end{smallmatrix}\big)$
\subsubsection{Identity Pose Consistency.}
Self-supervised monocular depth estimation operates on the underlying assumption of moving camera and previous works address this by proposing masking strategies to filter out pixels with no overall motion, either due to static camera or moving objects \cite{monodepth2}. This work, in addition, forces the pose network to explicitly inspect static images and estimate no relative ego-motion. This is done by giving the pose network same frames and having a loss that minimizes any ego-motion $\mathbf{T}_t^t = \big(\begin{smallmatrix}
 \mathbf{e}_t^t \\
 \mathbf{t}_t^t 
\end{smallmatrix}\big) \in \mathfrak{se}(3)$. The loss is as follows

\begin{equation}
\label{eq:id}
\mathcal{L}_{\mathbf{T}_{id}} = \lVert \mathbf{T}_t^t \rVert_1 = \lVert \mathbf{e}_{t}^t \rVert_1 + \lVert \mathbf{t}_{t}^t \rVert_1
\end{equation}

\subsubsection{Cycle Pose Consistency.}
\begin{figure}[hbt!]
 \centering
  \begin{tabular}{ccc}
  \includegraphics[width=0.38\textwidth]{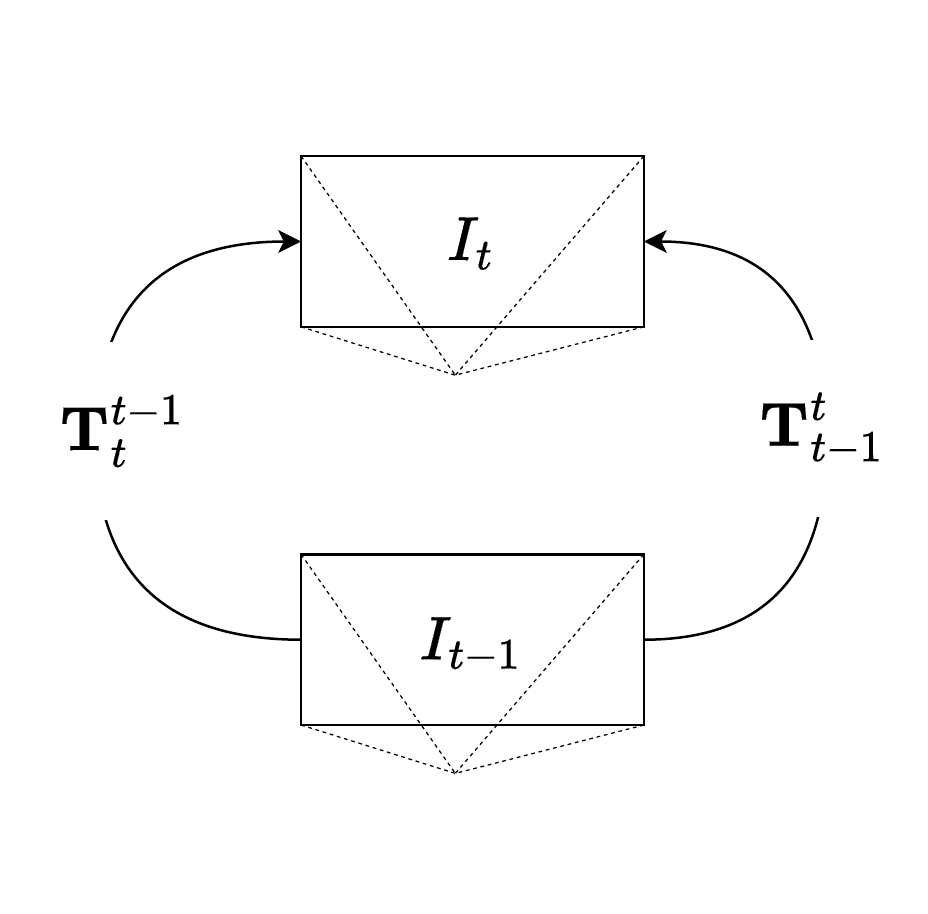} &
  \includegraphics[width=0.23\textwidth]{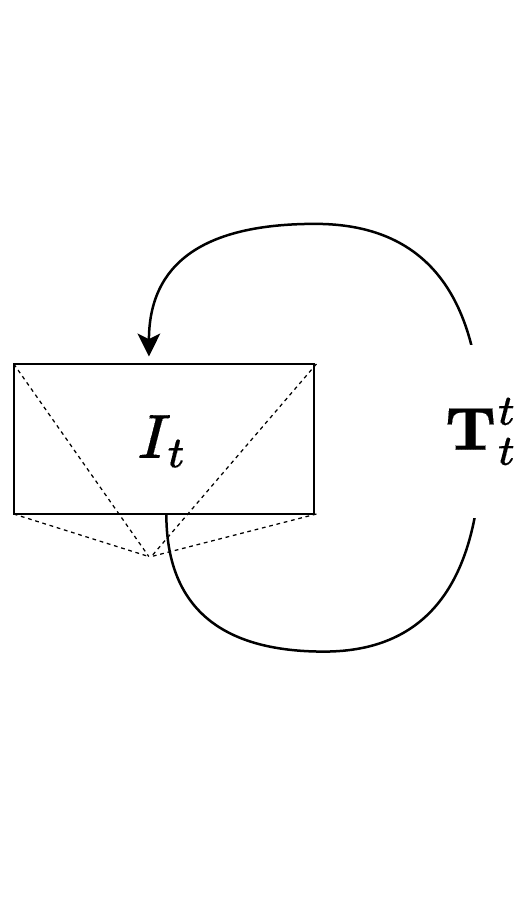} &
  \includegraphics[width=0.38\textwidth]{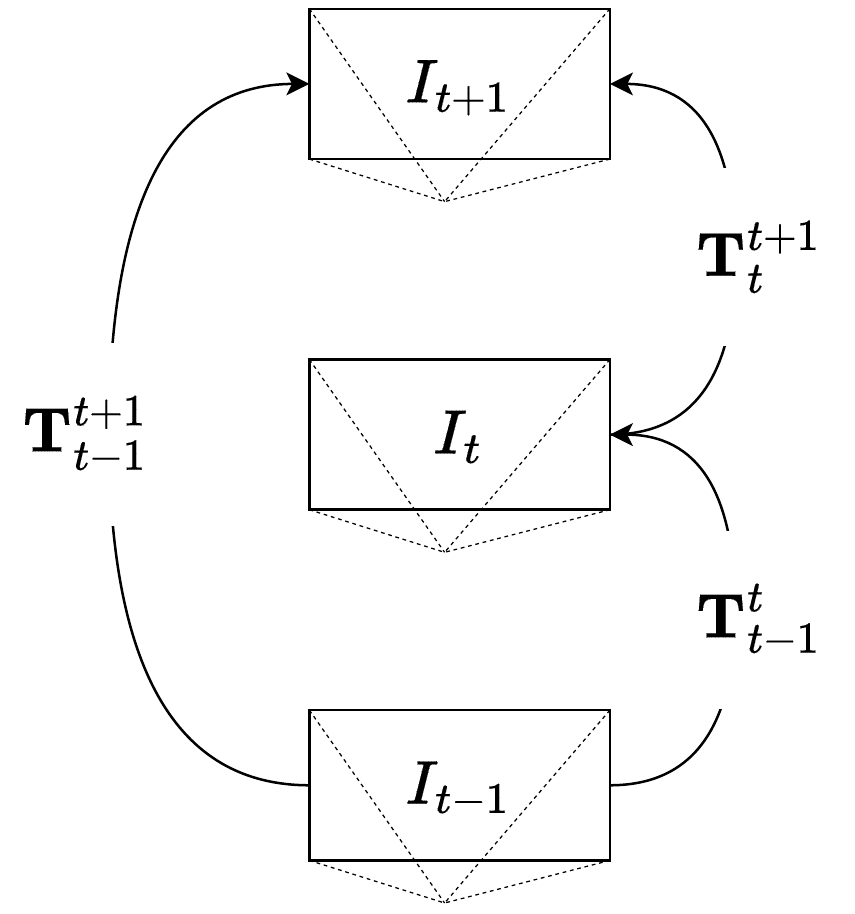}\\
  (a)&(b)&(c)
  \end{tabular}
  \caption{Illustration of the proposed: (a) forward-backward, (b) identity and (c) cycle, pose constraints}
  \label{fig:cyclic}
\end{figure}

Cyclic pose consistency states that the poses need to be consistent in a cyclic manner, \ie the combined pose from $t-n$ to $t+n$ via $t$, should be the same as the direct ego-motion between them, $\mathbf{T}^{t+n}_{t-n} \stackrel{!}{=} \mathbf{T}_t^{t+n}\times \mathbf{T}_{t-n}^{t}$, as illustrated in \cref{fig:cyclic}(c). The loss is parameterized as follows

\begin{align}
\label{eq:cyclic}
  \mathcal{L}_{\mathbf{T}_{cyc}} = d(\mathbf{T}^{t+n}_{t-n}, \ \mathbf{T}_t^{t+n}\times \mathbf{T}_{t-n}^{t})
\end{align}

\section{Experiments}
To maintain consistency and comparability, the exact experimentational setup of baseline \cite{monodepth2} is maintained. 
Following the established protocols by Eigen \etal \cite{eigen14}, we train and test on Zhou \etal \cite{sfmlearner}'s splits of the KITTI \cite{kitti} raw dataset, with a widely-used pre-processing to remove static frames \cite{sfmlearner} and cap the depth at $80$m. We evaluate the depth model using popularly used metrics from Eigen \etal \cite{eigen14}. We also consider the improved ground truth depth maps \cite{kittiimprovedsparse17} for evaluation, which uses stereo information and 5 consecutive frames to accumulate LiDAR points to handle moving objects, resulting in high quality depth maps. We weight our loss with a weight of $0.1$ when addition to the final objective.

\subsection{Temporal Scale Consistency of Predicted Depth}
The main goal of this work is to minimize scale inconsistencies. We introduce a simple metric to measure the scale consistency across consecutive frames


\begin{table*}[hbt!]
\centering
\resizebox{\columnwidth}{!}{\begin{tabular}{@{}ccccccccc@{}}
\hline
\toprule
 & \multicolumn{5}{c}{\cellcolor{red!25}{Lower is better}} & \multicolumn{3}{c}{\cellcolor{blue!25}{Accuracy (Higher is better)}} \\
\cmidrule(r){2-6}\cmidrule(l){7-9}
 Constraint & $\frac{\sigma(\text{scales})}{\mu(\text{scales})}$ & Abs Rel  &Sq Rel &RMSE &RMSE log &$\delta$ $<$ 1.25 & $\delta$ $<$ $1.25^{2}$ &$\delta$ $<$ $1.25^{3}$ \\
\bottomrule
\hline
 - &       0.096 & 0.116 &  0.896 &  4.915 &  0.194 &  0.872 &  0.957 &  0.981  \\
 $\mathcal{L}_{\mathbf{T}_{id}}$ & 0.090 & 0.116 &  0.897 &  4.823 &  0.193 &  0.874 &  0.959 &  0.981  \\
 $\mathcal{L}_{\mathbf{T}_{fb}}$ & 0.088 & 0.113 &  0.867 &  4.796 &  0.191 &  0.878 &  0.960 &  0.981  \\
 $\mathcal{L}_{\mathbf{T}_{cyc}}$ & 0.090 & 0.113 &  0.851 &  4.809 &  0.191 &  0.877 &  0.959 &  0.981  \\
 \midrule
 - &       - &  0.120 &  0.924 &  4.962 &  0.198 &  0.860 &  0.956 &  0.980 \\ 
 $\mathcal{L}_{\mathbf{T}_{fb}}$ & - &  0.116 &  0.885 &  4.855 &  0.195 &  0.867 &  0.958 &  0.981 \\ 
\bottomrule
\hline
\end{tabular}}
\caption{Depth evaluation with each of our constraints applied individually. The second column shows the coeffecient of correlation (normalized standard deviation) of scales as a measure of consistency. The top section uses ground-truth scaling in order to make the depth metrics comparable. In the bottom section, a per-sequence (in contrast to per-frame) median scaling is used. Our constraints show reduction in inconsistencies, while also slightly improving the depth.} 
\label{tab:ablation}
\end{table*}

\textbf{Consistency Metric.} We hypothesize that in practice, it would be possible to calculate the actual scale once and use it for the rest of the sequence. Thus, the actual value of the scale does not matter but only its normalized variation \wrt time. We define the coefficient of correlation of the computed scale factors as $ \frac{\sigma(\text{scales})}{\mu(\text{scales})},$ where scale=$\frac{\text{median}(D_{\text{true}})}{\text{median}(D_{\text{pred}})}$, as a measure of temporal scale consistency in depth. 

\cref{tab:ablation} compares our proposed constraints on the standard depth metrics as well as the proposed consistency metric. Additionally, for each sequence in the KITTI test set, we calculate the scale as the median of all per-frame scales and use it throughout the sequence, resulting in the evaluation shown in \cref{tab:ablation}'s bottom two rows. Our constraints show reduction in inconsistencies, while also slightly improving the depth.

\subsection{Depth Evaluation}
\cref{tab:comparison} compares our best model (with $\mathcal{L}_{\mathbf{T}_{cyc}}$) with works which tackle the scale ambiguity / scale consistency problem. Works which make use of some kind of ground truth supervision, shown in the top section of \cref{tab:comparison}, disambiguate scale as a byproduct and, as a result, are automatically scale consistent. 
Different types of supervision signals have been used as mentioned in the \emph{Supervision} column. In the bottom section, works tackling the problem without additional supervision are mentioned. Our work belongs in this category.

\begin{table*}
\centering
\resizebox{\columnwidth}{!}{\begin{tabular}{@{}ccccccccc@{}}
\hline
\toprule
& &\multicolumn{4}{c}{\cellcolor{red!25}{Error (Lower is better)}} &\multicolumn{3}{c}{\cellcolor{blue!25}{Accuracy (Higher is better)}} \\
\cmidrule(r){3-6}\cmidrule(l){7-9}
Method  & Supervision &Abs Rel  &Sq Rel &RMSE &RMSE log &$\delta$ $<$ 1.25 & $\delta$ $<$ $1.25^{2}$ &$\delta$ $<$ $1.25^{3}$ \\
\bottomrule
\hline
 Sparse-to-Cont \cite{sparse2dense19} & M+D & 0.118 & 0.630 & 4.520 & 0.209 & \textbf{0.898} & \textbf{0.966} & \textbf{0.985} \\
 DNet \cite{planetowards20} & M+h & 0.113 & 0.864 & 4.812 & 0.191 & 0.877 & 0.960 & 0.981 \\
 pRGBD-Refined \cite{prgbdslam20} & M+SLAM & 0.113 & 0.793 & 4.655 & \textbf{0.188} & 0.874 & 0.960 & 0.983 \\
 Kuznietsov \etal \cite{semi-supervised17} & M+D & 0.113 & \textbf{0.741} & \textbf{4.621} & 0.189 & 0.862 & 0.960 & 0.986 \\
 TrainFlow \cite{scalevo20} & M+SfM & 0.113 & 0.704 & 4.581 & 0.184 & 0.871 & 0.961 & 0.984 \\
 G2S R50 \cite{chawlavarma2021multimodal} & M+v & 0.112 & 0.894 & 4.852 & 0.192 & 0.877 & 0.958 & 0.981 \\
 PackNet-SfM \cite{packnet} (ResNet18) & M+v & \textbf{0.111} & 0.829 & 4.788 & 0.199 & 0.864 & 0.954 & 0.980 \\
  VA-Depth \cite{vadepth22} & M+h & 0.112 & 0.864 & 4.804 & 0.190 & 0.878 & 0.959 & 0.982 \\
\hline
 vid2depth \cite{vid2depth} & M & 0.163 & 1.240 & 6.220 & 0.250 & 0.762 & 0.916 & 0.968 \\
 DF-Net \cite{dfnet18}& M & 0.150 & 1.124 & 5.507 & 0.223 & 0.806 & 0.933 & 0.973 \\
 Sheng \etal \cite{keyframecyclewarp21} & M & 0.139 & 1.021 & 5.418 & 0.209 & 0.803 & 0.937 & 0.976 \\
 Li \etal \cite{li2021unsupervised} & M & 0.130 & 0.950 & 5.138 & 0.209 & 0.843 & 0.948 & 0.978 \\
 SC-SfMLearner \cite{scsfmlearner21} & M & 0.119 &	0.857 &	4.950 &	0.197 &	0.863 &	0.957 &	\underline{0.981} \\
 TC-Depth \cite{tcdepth21} (only $\mathcal{L}_{geo}$) & M & \underline{0.113} & 0.904 & \underline{4.773} & 0.193 & \underline{0.877} & \underline{0.959} & 0.980 \\
 Wang \etal \cite{3dscale21} & M & \textbf{0.109} & \textbf{0.779} & \textbf{4.641} & \textbf{0.186} & \textbf{0.883} & \textbf{0.962} & \textbf{0.982} \\
 Baseline (Monodepth2)~ \cite{monodepth2} & M  &0.115	&0.903	&4.863 	&0.193	&\underline{0.877}	&\underline{0.959}	& \underline{0.981} \\ 
\textbf{Ours} & M & \underline{0.113} &  \underline{0.851} & 4.809 &  \underline{0.191} &  \underline{0.877} &  \underline{0.959} &  \underline{0.981} \\
\bottomrule
\hline
\end{tabular}}
\caption{Comparison of our results with works that focus on temporal consistency, on the test set of KITTI \cite{kitti}'s Eigen \cite{eigen14} split. Works in the botom section use monocular (M) supervision, while those in the top section use additional supervision: D for LIDAR depths, h for camera extrinsics (height), v for velocity (GT pose) and SfM/SLAM for depth hints from classical methods. For fair comparison, only the contributions tackling the inconsistency problem are compared against. The best and second best methods are in bold and underlined respectively.} 
\label{tab:comparison}
\end{table*}

\cref{tab:comparison} shows that our proposed pose consistency constraint gives improved performance with respect to baseline \cite{monodepth2}, although this was not our goal, showing that temporal consistency is also important for the accuracy

\subsection{Ego-motion Evaluation}
Following Zhou \etal \cite{sfmlearner}'s protocols, we also evaluate our ego-motion network. We train on KITTI odometry splits sequences 0-8 and test on sequences 9 and 10. Similar to the related methods, we compute the absolute trajectory error (ATE) averaged over all overlapping 5-frame snippets. Since our pose network only takes 2 frames as input, we aggregate the relative poses to create 5-frame trajectories. \cref{tab:pose} summarizes our improvements with respect to our baseline \cite{monodepth2}. Our proposed loss constrains the solution space, thereby making a better optimum easier to achieve.

\begin{table*}[hbt!]
\centering
\begin{tabular}{@{}lcc@{}}
\hline
& Seq. 09 & Seq. 10 \\
\toprule
DF-Net \cite{dfnet18} & $0.017\pm 0.007$ & $0.015\pm 0.009$ \\
Wang \etal \cite{3dscale21} & $0.014\pm0.008$ & $0.014\pm0.010$ \\ 
Baseline \cite{monodepth2} & $0.017\pm0.008$ & $0.015\pm0.010$ \\
Ours ($\mathcal{L}_{\mathbf{T}_{cyc}}$) & $0.016\pm 0.008$ & $0.014\pm 0.010$ \\
\bottomrule
\hline
\end{tabular}
\caption{Ego-motion estimation results: average absolute trajectory error, and standard deviation, in meters, on KITTI Odommetry dataset \cite{kitti}. Trained on Seq. 0-8 and tested on Seq. 9 and 10. Our pose constraints show improvement with respect to baseline \cite{monodepth2}.} 
\label{tab:pose}
\end{table*}
\vspace{-1em}

\section{Conclusion and Future Work}
Self-supervised monocular depth and ego-motion estimation methods suffer not only from scale ambiguity but also from scale inconsistency. While including some kind of ground truth (GT) supervision not only disambiguates scale but also enforces scale consistency, it is not always plausible to have access to accurate ground-truth information. We propose ego-motion constraints that do not require any additional GT supervision. Via experimentation, we show that our proposed constraints not only decrease the inconsistencies but also improve the depth's and ego-motion's performance compared to baseline.

Our constraints do not aim to compete but complement the ones used in literature, for example, SC-SfMLearner \cite{scsfmlearner21}'s depth consistency. We have looked at how individual constraint performs. We leave the effect of their interactions with one another and constraints used by previous works for the future. We also do not loosen the static scene assumption. If the image motion is dominated by moving objects, it is indeed difficult to estimate the true ego-motion caused just by cameras. It would be interesting to generalize these constraints with denser motion models, such as Li \etal \cite{li2021unsupervised}'s piece-wise rigid flows considering dynamic objects or dense per-pixel optical flows \cite{draft22}.


\bibliographystyle{splncs04}
\bibliography{egbib}
\end{document}